%% file: neurips_2026.tex
\documentclass{article}
\usepackage{graphicx}
\usepackage{booktabs}
\usepackage{tabularx}
\usepackage{multirow}
\usepackage{array}
\usepackage{amsmath}
\usepackage{enumitem}
    \PassOptionsToPackage{numbers, compress}{natbib}

 \usepackage[dblblindworkshop, final]{neurips_2026}

\workshoptitle{Sim2Science: ML with Imperfect Scientific Models}

\usepackage[utf8]{inputenc} 
\usepackage[T1]{fontenc}    
\usepackage{hyperref}       
\usepackage{url}            
\usepackage{booktabs}       
\usepackage{amsfonts}       
\usepackage{nicefrac}       
\usepackage{microtype}      
\usepackage{xcolor}         

\newcommand{\pv}[1]{\textcolor{black}{#1}}

\title{Benchmarking Machine Learning for Chaotic Three-Body Dynamics in Astrophysics}

\author{%
  Sheng Wan \\
  École Polytechnique, IP Paris\\
  \texttt{sheng.wan@polytechnique.edu} \\
  \And
  Pavan Vynatheya \\
  \small{CITA, University of Toronto} \\
  \texttt{pavanvyn@cita.utoronto.ca} 
  \And
  Duo Xu \\
  \small{CITA, University of Toronto} \\
  \texttt{xuduo@cita.utoronto.ca} \\
  \And
  Li Kevin Wenliang \\
  Google DeepMind \\
  \texttt{kevinwli@outlook.com} \\
  \And
  Ye Zhu \\
  LIX, CNRS, École Polytechnique, IP Paris\\
  \texttt{ye.zhu@polytechnique.edu} \\
}

\begin{document}

\maketitle

\begin{abstract}
Machine learning (ML) and generative AI have recently made impressive progress in understanding physics.
We explore a hierarchy of prediction problems stemming from chaotic three-body systems and benchmark existing ML approaches, including general-purpose prediction algorithms and those that build on top of foundational physics models. 
We first simulate the trajectories of triple systems near the critical boundary of stable/unstable initial conditions. We then design a hierarchy of learning tasks, focusing mainly on two broad problems: stability prediction and trajectory prediction. The former classifies whether a given triple system has diverged or not up to a specific horizon, while the latter predicts the exact trajectory, both given suitable input features. We benchmark a range of ML approaches, from simple regression methods to state-of-the-art foundational physics models, across these tasks, and compare them against astrophysical methods such as the Mardling–Aarseth stability criterion. Our experiments reveal a clear order of task difficulties in chaotic three-body dynamics: early dynamics enable strong stability prediction, while precise long-horizon trajectory forecasting remains difficult.
Code and dataset are publicly available at \url{https://github.com/Simonwan03/ThreeBodyMLBench}.

\end{abstract}

\vspace{-0.05in}
\section{Introduction}
\label{sec:intro}
\vspace{-0.05in}

Recent research in machine learning has increasingly shifted from learning the semantics of data such as language and images ~\citep{devlin2019bert,radford2021clip} toward understanding the physics underlying everyday observations, as evidenced by the growing popularity of world models ~\cite{ha2018worldmodels, hafner2019dream, assran2025vjepa2} and recent physics-aware benchmarks~\cite{Motamed_2026_WACV,bansal2025videophy, meng2025worldsimulator}. 
In this work, we focus on one of the fundamental problems in physics that remains relatively underexplored in current AI research: the three-body problem~\cite{valtonen2006threebody,breen2020newton,lalande2022predicting,vynatheya2022algebraic}. It is a classical problem for understanding the stability and dynamics of stellar systems, and provides one of the simplest yet best-controlled settings for benchmarking ML methods.

A triple system consists of three masses whose motions are deterministically governed by their initial physical conditions through Newtonian dynamics.
Their long-term stability and orbital evolution are relevant to a broad range of astrophysical phenomena~\cite{toonen2016evolution}.
Depending on the initial conditions, a triple system may either remain stable over long timescales or rapidly diverge and become unbounded. However, between these two regimes lies a critical region in which systems that only differ very slightly in initial conditions can lead to markedly different stability outcomes~\cite{tory2022empirical,vynatheya2022algebraic}. 
We focus on this boundary regime from which we select the initial conditions for our simulations, used as the ground-truth trajectories.

\begin{figure}[t]
\centering

\begin{minipage}[c]{0.4\linewidth}
\centering
\scriptsize
\renewcommand{\arraystretch}{1.0}
\setlength{\tabcolsep}{3pt}

\begin{tabularx}{\linewidth}{l c X}
\toprule
\textbf{Feature} & \textbf{\# Ch.} & \textbf{Physical meaning} \\
\midrule
\texttt{aratios} & 1 & semi-major-axis ratio $a_{\rm in}/a_{\rm out}$ \\
\texttt{mratios} & 2 & mass ratios of the hierarchical triple \\
\texttt{eccs} & 2 & eccentricities $e_*$ of the orbits \\
\texttt{incs} & 2 & inclinations of the orbital planes $i_*$ \\
\texttt{APs} & 4 & arguments of periapsis $\omega_{*}$ \\
\texttt{LANs} & 4 & longitudes of ascending node $\Omega_{*}$\\
\bottomrule
\end{tabularx}
\end{minipage}
\hfill
\begin{minipage}[c]{0.59\linewidth}
\centering
\includegraphics[
    width=\linewidth,
]{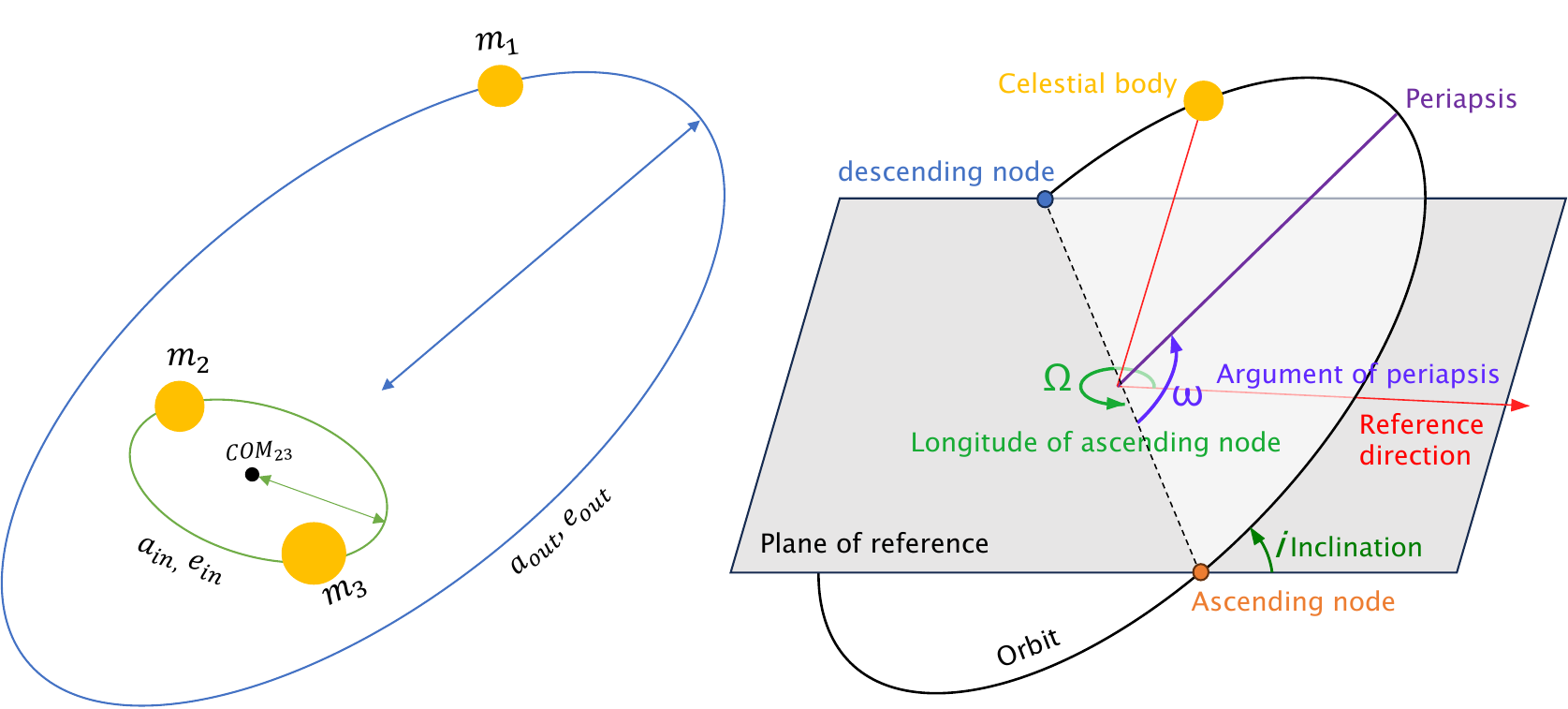}
\end{minipage}

\caption{
Left: Orbital representation used in our experiments. $*$ stands for either the inner or the outer orbit. Right: Illustration of orbital features. Adapted from Lasunncty~\cite{lasunncty2007orbit}, with changes to colors and annotations. The original and this adapted figure are licensed under \href{https://creativecommons.org/licenses/by-sa/3.0/}{CC BY-SA 3.0}.
}
\label{fig:orbital_representation}
\vspace{-0.2in}
\end{figure}

Given the trajectories, we design a suite of tasks with different levels of difficulty, broadly categorized as stability prediction and trajectory prediction. Stability prediction aims to determine whether or not a given triple system will remain bound or eventually become unbounded. The input features are either its physical initial conditions or a short prefix of the ground-truth trajectory ~\cite{lalande2022predicting}. Trajectory prediction goes beyond the binary stability outcome and predicts the detailed dynamical evolution of the system. 
To see how hard these tasks are, we benchmark a range of ML approaches against conventional astrophysical methods, spanning from classical supervised learning models~\cite{rumelhart1986learning,lecun1998gradient,bai2018empirical,
chen2016xgboost,cortes1995support} to more recent foundation models~\cite{lai2026panda, ansari2025chronos2, podest2026tirex, khwaja2026toto} and our own fine-tuned variants. We additionally consider established analytical stability criteria in astrophysics: Mardling–Aarseth~\cite{mardling2001tidal} and Vynatheya et al. criterion~\cite{vynatheya2022algebraic}. Our extensive experiments show that the performance of ML methods varies depending on the task specification; while short-term and coarse-grained properties (e.g. stability) can be learned with high accuracy, long-term trajectory prediction remains substantially more challenging.
Overall, our main contributions are summarized as follows: 

\begin{itemize}
    \item We construct a large-scale controlled perturbation-ensemble dataset of chaotic hierarchical three-body systems concentrated near the challenging astrophysical stability boundary.
    \item We design a hierarchical suite of stability and trajectory prediction tasks that probe different levels of dynamical prediction difficulty, and evaluate a broad range of ML approaches against conventional astrophysical methods.
    \item Through systematic experiments and analysis, we characterize both the capabilities and limitations of ML in learning chaotic three-body dynamics, providing insights into the limitations and challenges of existing ML approaches.
\end{itemize}

\vspace{-0.1in}
\section{Related Work}
\label{sec:related}
\vspace{-0.1in}

\textbf{Three-body systems with boundary conditions.}
Three-body systems have long been studied for orbital stability. Traditional approaches rely primarily on analytical or semi-empirical criteria, such as the Eggleton--Kiseleva and Mardling--Aarseth criteria~\cite{eggleton1995empirical,mardling2001tidal}, which characterize stability using orbital elements and mass ratios. Although computationally efficient and widely used, these criteria can become less reliable near stability boundaries or outside the regimes in which they were calibrated~\citep{tory2022empirical,vynatheya2022algebraic}. ML methods have been applied to long-term stability prediction in both compact planetary systems and hierarchical triples. Tamayo et al.~\cite{tamayo2020predicting} combine short \(N\)-body integrations with physically motivated summary statistics to predict stability over up to \(10^{9}\) orbits. Neural networks have been used to predict instability timescales~\cite{cranmer2021bayesian} and hierarchical triple-star stability~\cite{vynatheya2022algebraic,vynatheya2023quadruple}.
On trajectory prediction, Breen et al.~\cite{breen2020newton} used neural networks to approximate chaotic three-body trajectories over bounded time intervals, targeting efficient trajectory prediction rather than long-term stability characterization.
We evaluate both stability and trajectory predictions within a unified hierarchy of tasks,
which allows us to systematically study how predictive performance changes with the available dynamical information and prediction horizon in strongly chaotic regimes.


\textbf{Physics-focused benchmarks for machine learning.}
For broader context, physics-focused benchmarks have increasingly been used to evaluate whether ML models can simulate or predict physical dynamics. For instance, PDEBench~\cite{takamoto2022pdebench}, LagrangeBench~\cite{toshev2023lagrangebench}, and The Well~\cite{ohana2024well} provide standardized simulation datasets for evaluating surrogate models across diverse physical systems, with emphasis on rollout accuracy, generalization, and physically meaningful errors. Complementary benchmarks for world models and intuitive physics, including Physion~\cite{bear2021physion}, Physics-IQ~\cite{Motamed_2026_WACV}, and WorldModelBench~\cite{li2025worldmodelbench}, examine whether learned models capture physical principles rather than merely producing visually plausible visual predictions. 
Discrete chaotic systems, such as elementary cellular automata~\citep{burtsev2024learning, zhang2025intelligence}, have also been used as benchmarks for studying the capabilities of neural architectures.
Our benchmark complements them by providing a simple but well-controlled setting for evaluating how the predictive performance is influenced by the targets and input features of carefully constructed tasks all related by the three-body dynamics.

\vspace{-0.1in}
\section{Astrophysical Setup and Learning Tasks}
\label{sec:dataset}

\vspace{-0.05in}
\subsection{Astrophysical Setup and Dataset Construction}
\label{subsec:simulation_preprocessing}
\vspace{-0.05in}

\textbf{Hierarchical triple systems and controlled perturbations.}
Hierarchical triple systems provide a canonical setting for studying the long-term dynamics and stability of the three-body problem. We construct a controlled perturbation-ensemble dataset comprising \(5{,}000\) \emph{base \pv{triple} ensembles of systems}. \pv{The initial physical conditions are close to the critical stability boundary, predicted by the MLP model from Vynatheya et al.~\cite{vynatheya2022algebraic,vynatheya2023quadruple}, in phase space. Among these, roughly 40\% of the systems are predicted stable by this model.} For each base triple system, we \pv{generate} \(32\) \pv{variations} from the same base initial conditions and a prescribed perturbation of selected initial-condition variables, yielding \(160{,}000\) \emph{individual triple systems} in total. \pv{We then employ the Newtonian \textit{N}-body code \texttt{REBOUND}~\cite{rein2012rebound}, using the 15th-order integrator \texttt{IAS15}~\cite{rein2015ias15}, to compute the gravitational dynamics and the resulting trajectories.} Each trajectory is sampled every 5 outer orbits for a total of 10,000 outer orbits, yielding 2,001 time points.

 \pv{The \(32\) variations associated with one base system differ only in orbital phases or slight semi-major-axis perturbations, and thus should not be treated as independent samples}. We therefore treat each 32-realization perturbation ensemble as the fundamental unit for dataset splitting and ensemble-level evaluation. This design preserves the response of each system to controlled changes in its initial conditions and provides an empirical distribution of stability outcomes, enabling both individual-level and ensemble-level tasks, such as stable-fraction estimation and survival curve analysis. The full perturbation schedule, parameter ranges, integration settings, and preprocessing procedure are described in Appendix~\ref{app:ic_variations} and~\ref{app:dataset}.

\textbf{Scale-invariant processing.}
We represent the dynamical state using dimensionless orbital descriptors rather than raw Cartesian positions and velocities. This choice reduces sensitivity to nuisance degrees of freedom and provides a representation more directly aligned with the underlying orbital dynamics.
These descriptors also provide a physically meaningful representation for the prediction problems considered in this work.
%
Unless otherwise specified, all experiments use this orbital representation. The resulting \texttt{orbital15} feature set, summarized in Fig.~\ref{fig:orbital_representation}, provides a common state representation across stability and trajectory prediction tasks, enabling consistent comparisons across model families.

\vspace{-0.05in}
\subsection{Stability of Triple-Star Systems}
\vspace{-0.05in}


The overall goal of the stability prediction task is to determine whether a given triple-star system will remain bounded at a future timestamp, typically far from its current state. Here, we focus on several sub-tasks that predict stability based on different types of input information, as well as a survival prediction task under controlled small perturbations.
In particular, we use each base system, which consists of 32 realizations as detailed in Sec.~\ref{subsec:simulation_preprocessing}, as the minimal unit for prediction tasks.

\textbf{\pv{Task A:} Stability prediction from initial physical conditions.}
Given the initial condition $\mathbf{c}_i$ of a nominal system $i$, we seek to predict its long-term stability at a future horizon $H$. Because each nominal system is simulated with $V$ nearby phase-space perturbations, we define its stability as the fraction of variations that remain bound until $H$:
{\footnotesize
\setlength{\abovedisplayskip}{3pt}
\setlength{\belowdisplayskip}{3pt}
\begin{equation}
    \widehat{q}_i(H)
    = f_{\mathrm{A}}(\mathbf{c}_i),
    \qquad
    q_i(H)
    = \frac{1}{V}\sum_{j=1}^{V}
      \mathbb{I}\!\left(T_{ij} \geq H\right),
    \label{eq:task_a}
\end{equation}
}
$f_{\mathrm{A}}$ denotes the prediction model, $T_{ij}$ is the instability time of the $j$-th variation of ensemble $i$, and $\mathbb{I}(\cdot)$ is the indicator function. Thus, $q_i(H)\in[0,1]$ is the empirical stable fraction and $\widehat{q}_i(H)$ is its prediction. For the binary all-stable setting, the label is defined as $y_i=\mathbb{I}\!\left(q_i(H)=1\right)$, $y_i=1$ indicates that all variations of an ensemble remain stable and $y_i=0$ indicates that at least one variation diverges. This initial-condition-based formulation is closely related to conventional analytical stability criteria~\cite{mardling2001tidal} and allows rapid screening without performing a long-term numerical integration.

\textbf{\pv{Task B:} Stability prediction from a short-term trajectory prefix.}
Given a short-term trajectory prefix
$\mathbf{X}_{ij}^{(P)}
 = [\mathbf{x}_{ij}(t_1),\ldots,\mathbf{x}_{ij}(t_P)]
 \in \mathbb{R}^{P\times d}$
for variations $j$ of ensemble of system $i$, we seek to predict whether the individual system remains stable until a future horizon $H$, where $t_P < H$. The task is formulated as:
{\footnotesize
\setlength{\abovedisplayskip}{3pt}
\setlength{\belowdisplayskip}{3pt}
\begin{equation}
    \widehat{p}_{ij}(H)
    = f_{\mathrm{B}}\!\left(\mathbf{X}_{ij}^{(P)}\right),
    \qquad
    y_{ij}(H)
    = \mathbb{I}\!\left(T_{ij} \geq H\right).
    \label{eq:task_b}
\end{equation}
}Here, $\mathbf{x}_{ij}(t_p)\in\mathbb{R}^{d}$ is the $d$-dimensional orbital state observed at timestamp $t_p$, $P$ is the prefix length, and $f_{\mathrm{B}}$ outputs the predicted probability $\widehat{p}_{ij}(H)$ of survival. The binary label $y_{ij}(H)$ equals one for a stable individual and zero for an individual system that becomes unstable before $H$. Unlike Task A, this formulation exploits early dynamical evolution, which may reveal precursors of instability that are not apparent from the initial conditions alone \cite{tamayo2020predicting}.

\textbf{\pv{Task C:} Survival prediction within an ensemble of systems.}
Given the collection of short-term prefixes
$\mathcal{X}_i=\{\mathbf{X}_{ij}^{(P)}\}_{j=1}^{V}$
from all $V$ variations of ensemble $i$, we seek to predict the survival probability of each variation at multiple future horizons
$H_1<\cdots<H_K$. We parameterize the prediction using discrete conditional hazards:
{\footnotesize
\setlength{\abovedisplayskip}{3pt}
\setlength{\belowdisplayskip}{3pt}
\begin{equation}
\begin{aligned}
    \widehat{h}_{ij,k}
    &=
    \Pr\!\left(
        H_{k-1} \leq T_{ij} < H_k
        \mid T_{ij} \geq H_{k-1},\,\mathcal{X}_i
    \right), \\
    \widehat{S}_{ij}(H_k)
    &=
    \prod_{\ell=1}^{k}
    \left(1-\widehat{h}_{ij,\ell}\right),
    \qquad
    \widehat{Q}_i(H_k)
    =
    \frac{1}{V}\sum_{j=1}^{V}
    \widehat{S}_{ij}(H_k).
\end{aligned}
\label{eq:task_c}
\end{equation}
}
$\widehat{h}_{ij,k}$ is the predicted conditional probability that variation $j$ becomes unstable during the interval $[H_{k-1},H_k)$, while $\widehat{S}_{ij}(H_k)$ is its predicted probability of remaining stable through $H_k$. $\widehat{Q}_i(H_k)$ denotes the predicted surviving fraction within ensemble $i$, $K$ is the number of prediction horizons, and $V=32$ in our experiments. The corresponding empirical target is
$Q_i(H_k)=\frac{1}{V}\sum_{j=1}^{V}\mathbb{I}(T_{ij} \geq H_k)$.
This formulation produces a monotonic survival curve rather than a single binary outcome, while naturally representing trajectories that remain stable throughout the finite simulation horizon.

\vspace{-0.05in}
\subsection{Trajectory Forecasting of Detailed Dynamics}
\vspace{-0.05in}


In addition to the stability of the ensemble of systems, we also investigate whether the existing ML methods can predict the fine-grained orbital evolution of chaotic three-body systems, characterized by a series of trajectory physical quantities. Specifically, we evaluate whether state-of-the-art pretrained time-series models can predict the future orbital evolution of chaotic three-body systems from a finite observation history. Given a 512-step context represented by \texttt{orbital15} features as mentioned in Fig.~\ref{fig:orbital_representation}, the model predicts the subsequent dynamics. We consider both a one-shot 128-step forecast and a longer 512-step open-loop rollout obtained by recursively applying the same predictor over four consecutive 128-step blocks without access to future observations. As orbital evolution is conventionally studied through numerical integration~\cite{valtonen2006threebody,rein2015ias15}, this task provides a natural test of whether pretrained time-series models can serve as data-driven predictors of chaotic dynamics.

\vspace{-0.05in}
\section{Experiments and Analysis}
\label{sec:exp}
\vspace{-0.05in}

\subsection{Experimental Setup}
\vspace{-0.05in}

\textbf{Baseline methods.}
The methods used in our experiments can be mainly classified into three categories, spanning domain-specific astrophysical criteria, task-specific supervised learning models, and recent pretrained time-series foundation models.
\textbf{1. Physics-based empirical criteria:} We include classical stability criteria from celestial mechanics, such as the Mardling--Aarseth criterion~\cite{mardling2001tidal} and Vynatheya et al. criterion~\cite{vynatheya2022algebraic}, which estimate the stability of hierarchical three-body systems using analytically motivated empirical relations between orbital and physical parameters.
\textbf{2. Task-specific supervised learning models:} We evaluate a range of standard machine learning and deep learning approaches trained specifically for each prediction task. These include feature-based models such as XGBoost, as well as neural architectures including multilayer perceptrons (MLP), convolutional neural networks (CNN), temporal convolutional networks (TCN), and attention-based models. Depending on the task, these models operate either on engineered dynamical summaries (\ref{app:dynamical_summary}) or directly on orbital trajectories. 
\textbf{3. Pretrained time-series foundation models:}
We further evaluate recent large-scale time-series models pretrained on diverse temporal data, as well as our fine-tuned variants. In particular, we consider models capable of multivariate sequence modeling and assess their ability to transfer general temporal representations to chaotic three-body dynamics, both directly and, where applicable, through task-specific adaptation.

\textbf{Evaluation metrics.}
We evaluate classification using accuracy, F1, AUROC, and AUPRC;
regression using MAE, MSE, RMSE, $R^2$, and correlation;
survival using horizon-wise MAE and IBS;
and forecasting using standardized MSE/MAE, correlation,
unit-circle error, and rollout error accumulation.
See Appendices~\ref{app:data_splits}
and~\ref{app:evaluation_metrics} for data splits
and metric definitions.

\vspace{-0.05in}
\subsection{Main Results and Analysis}
\vspace{-0.05in}
\begin{table}[t]
\centering
\caption{
Representative results across the benchmark.
Full results are reported in Appendix~\ref{app:result}.
}
\label{tab:main_results}
\small
\setlength{\tabcolsep}{3.2pt}
\resizebox{\linewidth}{!}{
\begin{tabular}{l l l l}
\toprule
Prediction setting & Representative method & Key result & Main finding \\
\midrule

Initial-condition stability &
XGBoost &
Binary: AUROC $0.8525$, Fraction: MAE $0.2330$ &
Limited predictability \\

Prefix-based stability &
Dynamical summaries + XGBoost &
Binary: AUROC $0.9851$, Fraction: MAE $0.0398$ &
Early dynamics improve predictability \\

Multi-horizon survival &
Chronos-2 + direct pooling &
MAE $0.0376$, IBS $0.0416$ &
Strong survival estimation \\

Trajectory forecasting &
Chronos-2 (zero-shot) &
MAE $0.228$ &
Competitive short-term forecasting \\
\bottomrule
\end{tabular}}
\vspace{-0.2in}
\end{table}

\textbf{Stability prediction from initial physical conditions.}
Initial physical conditions provide substantial but incomplete information about long-term stability. 
For binary classification, learned models consistently outperform the analytical stability criteria. 
XGBoost achieves the highest accuracy 0.805 and F1 score 0.535, compared with 0.619 accuracy and 0.441 F1 for the Mardling--Aarseth criterion~\cite{mardling2001tidal}. 
Meanwhile, the MLP classifier from Vynatheya et al. ~\cite{vynatheya2022algebraic} achieves the strongest threshold-independent discrimination, with an AUROC of 0.909 and an AUPRC of 0.718. 
For stable-fraction regression, XGBoost obtains the lowest MSE 0.088 and highest $R^2$ 0.444, while the MLP classifier from Vynatheya et al. ~\cite{vynatheya2022algebraic} achieves a slightly lower MAE 0.218 vs. XGBoost 0.233. Overall, these results show that initial orbital parameters contain meaningful predictive information beyond fixed analytical criteria, but still leave substantial uncertainty about long-term outcomes.

\textbf{Early dynamics are highly informative for stability prediction.}
Short trajectory prefixes contain strong signals of future stability. Using dynamical summary statistics, XGBoost achieves AUROCs of 0.965 and 0.985 with 96- and 512-step prefixes, respectively, consistently outperforming models trained directly on raw sequences. Frozen PANDA~\cite{lai2026panda} and Chronos-2~\cite{ansari2025chronos2} representations are also competitive, with Chronos-2 reaching an AUROC of 0.979 and an AUPRC of 0.955 at 512 steps. For the more demanding multi-horizon survival task, pretrained encoders achieve the strongest performance: the best Chronos-2 model obtains an overall MAE of 0.0376 and an IBS of 0.0416, compared with 0.310 and 0.132 for the train-mean baseline. These results suggest that early dynamical evolution contains strong precursors of long-term instability, while the most effective representation depends on the prediction objective: physically motivated dynamical summaries excel at binary classification, whereas pretrained sequence representations are particularly effective for multi-horizon survival prediction.

\textbf{Long-horizon trajectory forecasting remains challenging.}
Zero-shot foundation models already capture substantial dynamical structure in three-body trajectories, achieving performance comparable to the supervised multivariate iTransformer~\cite{liu2024itransformer} despite no task-specific training. For example, PANDA~\cite{lai2026panda} obtains an MSE of \(0.222\), close to the supervised iTransformer at \(0.212\), while Chronos-2~\cite{ansari2025chronos2} achieves the lowest MAE \(0.228\)  and unit-circle error \(0.227\). This suggests that pretrained temporal representations can transfer effectively to chaotic orbital dynamics. Nevertheless, prediction errors accumulate consistently during the 512-step open-loop rollout across all models. Such error accumulation represents a key limitation of current learning-based forecasters and becomes particularly severe in chaotic systems, where small prediction errors are progressively amplified over time. 
Thus, while foundation models exhibit encouraging zero-shot dynamical prediction ability, reliable fine-grained forecasting over long horizons remains an open challenge.

\vspace{-0.1in}
\section{Conclusion and Discussion}
\label{sec:conclu}
\vspace{-0.1in}

We introduced a benchmark for machine learning on chaotic three-body dynamics near the astrophysical stability boundary. Our results show that stability becomes more predictable with early dynamical observations, while physically motivated features and pretrained representations remain competitive. For trajectory forecasting, foundation models can match supervised models, but long-horizon prediction remains challenging.

\newpage

\section*{Acknowledgements}
This work is supported by the French National Research Agency (ANR)
via the “MiMoGC” JCJC grant (ANR-26-CE23-7469), and State funding managed by the French National Research Agency (ANR) under the France 2030 program, reference ANR-23-IACL-0005. 
D.X. acknowledges the support of the Natural Sciences and Engineering Research Council of Canada (NSERC), [funding reference number 568580]. D.X. also acknowledges support from the Eric and Wendy Schmidt AI in Science Postdoctoral Fellowship Program, a program of Schmidt Sciences.

{
    \small
    \bibliographystyle{ieeenat_fullname}
    \bibliography{main}
}

\clearpage
\input{checklist}
\clearpage

\appendix

\section*{Technical appendices and supplementary material}

\label{app:main_appendix}

\section{Construction of Initial Conditions and Variations}
\label{app:ic_variations}

\pv{The initial conditions have 32 phase-space variations of 5,000 base triple systems. Each triple consists of 3 point masses present in two nested binary orbits. The `inner' binary consists of the first two point masses, and the `outer' binary consists of the third point mass and the total inner binary mass. The initial physical parameter distributions are as follows:}
\begin{itemize}
    \item \pv{Three masses $m_1$, $m_2$, and $m_3$ are uniformly sampled, spanning 2 orders of magnitude. This results in two mass ratios $m_1/m_2$ ($\leq 1$) and $m_3/(m_1+m_2)$.}
    \item \pv{Two semimajor axes $a_{\rm in}$ and $a_{\rm out}$ are uniformly sampled, spanning 4 orders of magnitude. This results in one semimajor axis ratio $a_{\rm in}/a_{\rm out}$ ($< 1$).}
    \item \pv{Two eccentricities $e_{\rm in}$ and $e_{\rm out}$ are uniformly sampled between 0.01 and 0.99. To ensure an initially non-crossing hierarchical configuration, we reject systems satisfying: $a_{\rm in} (1 + e_{\rm in}) > a_{\rm out} (1 - e_{\rm out})$}.
    \item \pv{Six orbital angles are isotropically sampled: two inclinations $i_{\rm in}$ and $i_{\rm out}$ (range from $0$ to $\pi$ and uniform in their cosines), longitudes of ascending node $\Omega_{\rm in}$ and $\Omega_{\rm out}$ (uniform from $0$ to $2 \pi$), and two arguments of periapsis $\omega_{\rm in}$ and $\omega_{\rm out}$ (uniform from $0$ to $2 \pi$).}
\end{itemize}

\pv{In addition to these parameters, two true anomalies (orbital phases) $\nu_{\rm in}$ and $\nu_{\rm out}$ need to be specified. We first construct a $4\times4$ grid of variations of mean anomalies $M_{\rm in}$ and $M_{\rm out}$, keeping all the other parameters fixed,}
\pv{\[
M_{\rm in},M_{\rm out}
\in
\left\{
0,\frac{\pi}{2},\pi,\frac{3\pi}{2}
\right\}.
\]}
\pv{We then convert the mean anomalies to true anomalies by numerically solving Kepler's equation. Mean and true anomalies are identical for circular orbits but not for eccentric orbits.}

\pv{Finally, we consider another variation per system by perturbing the nominal inner semimajor axis ratios,}
\pv{\[
a_{\rm in}^{\rm pert}
=
a_{\rm in}^{\rm nominal}+10^{-6},
\]}
\pv{while all other orbital parameters and masses are held fixed.}

\pv{This results in a total of $2\times4\times4=32$ variations per base triple system (see also Table~\ref{tab:variation_construction}). Using zero-based indexing, the variation index is}
\pv{\[
v = 16s + 4p_{\rm in} + p_{\rm out},
\]}
\pv{where $s\in\{0,1\}$ denotes the nominal or perturbed inner-semimajor-axis
setting, and $p_{\rm in},p_{\rm out}\in\{0,1,2,3\}$ index the nominal
orbital phases. Consequently, variations $v=0,\ldots,15$ form the nominal
$4\times4$ phase grid, while variations $v=16,\ldots,31$ repeat the same
phase configurations after perturbing the inner semimajor axis.}

\pv{This construction serves two complementary purposes. The $4\times4$ phase
grid captures the dependence of the long-term dynamics on orbital phase,
while the paired semimajor-axis perturbations provide controlled probes of
sensitivity to extremely small changes in the initial conditions. In a
chaotic regime, such nearly identical initial configurations can evolve
toward substantially different long-term outcomes.}

\begin{table}[htbp]
\centering
\small
\caption{
Construction of the 32 initial-condition variations for each hierarchical
triple. Variations 1--16 span a $4\times4$ grid of nominal inner- and
outer-orbit phases. Variations 17--32 repeat the same phase grid after
increasing the inner semimajor axis by $10^{-6}$.
}
\label{tab:variation_construction}
\begin{tabular}{cccc}
\toprule
Nominal $j$ &
Perturbed $j$ &
$M_{\rm in}$ &
$M_{\rm out}$ \\
\midrule
1  & 17 & $0$          & $0$ \\
2  & 18 & $0$          & $\pi/2$ \\
3  & 19 & $0$          & $\pi$ \\
4  & 20 & $0$          & $3\pi/2$ \\
5  & 21 & $\pi/2$      & $0$ \\
6  & 22 & $\pi/2$      & $\pi/2$ \\
7  & 23 & $\pi/2$      & $\pi$ \\
8  & 24 & $\pi/2$      & $3\pi/2$ \\
9  & 25 & $\pi$        & $0$ \\
10 & 26 & $\pi$        & $\pi/2$ \\
11 & 27 & $\pi$        & $\pi$ \\
12 & 28 & $\pi$        & $3\pi/2$ \\
13 & 29 & $3\pi/2$     & $0$ \\
14 & 30 & $3\pi/2$     & $\pi/2$ \\
15 & 31 & $3\pi/2$     & $\pi$ \\
16 & 32 & $3\pi/2$     & $3\pi/2$ \\
\bottomrule
\end{tabular}
\end{table}

\section{\pv{Datasets and Statistics}}
\label{app:dataset}

\pv{Of the total 160,000 individual triple systems generated, we ignore 64 of them since their $N$-body simulations do not reach completion within our maximum time limit of 5 hr per system.}

\begin{figure}[!ht]
    \centering
    \includegraphics[width=\linewidth]{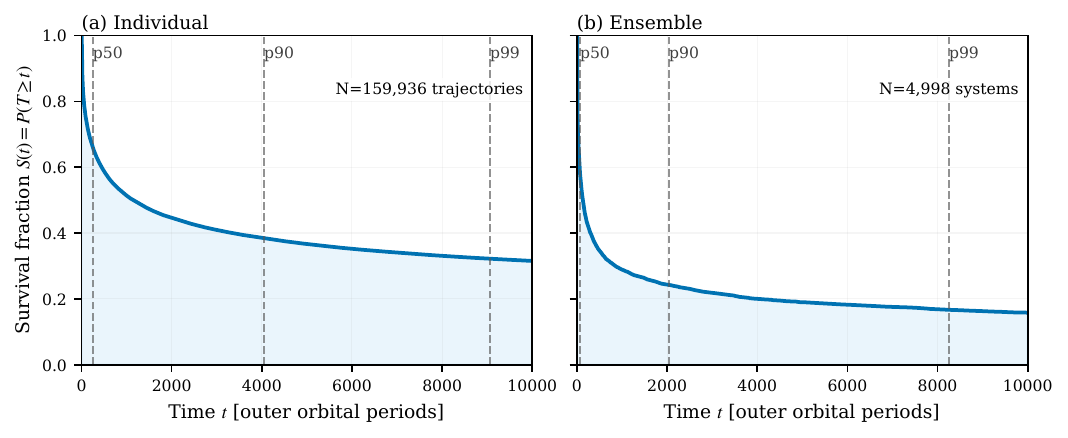}
    \caption{Distribution of the dataset at the individual and ensemble levels.}
    \label{fig:data_distribution}
\end{figure}

The dataset we offer has the following characteristics:
\begin{itemize}
    \item The full features included in the initial condition and evolving trajectory are listed in Table~\ref{tab:initial_condition_fields} and Table~\ref{tab:trajectory_fields}.
    \item Each ensemble of systems has 32 perturbations near the stability boundary that share similar initial conditions. The exact differences are shown in Appendix~\ref{app:ic_variations}.
    \item It has more systems near astrophysical stability boundary, and the detailed distribution is shown in Figure~\ref{fig:data_distribution}.
\end{itemize}

\begin{table*}[t]
\centering
\small
\caption{
Fields describing the initial conditions of the hierarchical triple systems.
}
\label{tab:initial_condition_fields}
\begin{tabularx}{\textwidth}{
    >{\ttfamily}p{3.2cm}
    X
}
\toprule
\rmfamily Key &
Physical meaning \\
\midrule

aratios &
Ratio of the inner to outer semimajor axes,
$a_{\rm in}/a_{\rm out}$, characterizing the hierarchical structure of the system. \\

mratios &
Two mass ratios, $m_1/m_2$ and $m_3/(m_1+m_2)$. The former describes the
mass ratio of the inner binary, while the latter measures the mass of the
outer body relative to the total mass of the inner binary. \\

eccs &
Eccentricities of the inner and outer orbits,
$(e_{\rm in}, e_{\rm out})$. \\

incs &
Inclinations of the inner and outer orbits with respect to the reference plane,
$(i_{\rm in}, i_{\rm out})$. These are not the mutual inclination; the mutual
inclination must be computed from the angular-momentum directions of the two orbits. \\

APs &
Arguments of periapsis of the inner and outer orbits,
$(\omega_{\rm in}, \omega_{\rm out})$. \\

LANs &
Longitudes of ascending node of the inner and outer orbits,
$(\Omega_{\rm in}, \Omega_{\rm out})$. \\



tAnos &
True anomalies of the inner and outer orbits,
$(\nu_{\rm in}, \nu_{\rm out})$, describing the initial orbital phases. \\









\bottomrule
\end{tabularx}
\end{table*}

\begin{table*}[htbp]
\centering
\footnotesize
\caption{
Fields storing the time evolution of the simulated hierarchical triple systems.
}
\label{tab:trajectory_fields}
\begin{tabularx}{\textwidth}{
    >{\ttfamily}p{4.0cm}
    X
}
\toprule
\rmfamily Key &
Physical meaning \\
\midrule

times\_complete &
Time grid for each trajectory. The dataset spans from $t=0$ to $10{,}000$ initial outer-orbital periods. \\

aratios\_complete &
Time evolution of the instantaneous osculating semimajor-axis ratio,
$a_{\rm in}(t)/a_{\rm out}(t)$. \\

mratios\_complete &
Time series of the mass ratios $m_1/m_2$ and $m_3/(m_1+m_2)$.
For point-mass integrations without mass transfer, these quantities remain constant. \\

eccs\_complete &
Instantaneous osculating eccentricities of the inner and outer orbits,
$(e_{\rm in}(t),e_{\rm out}(t))$. \\

incs\_complete &
Instantaneous inclinations of the inner and outer orbits with respect to
the reference plane, $(i_{\rm in}(t),i_{\rm out}(t))$. \\

APs\_complete &
Instantaneous arguments of periapsis of the inner and outer orbits,
$(\omega_{\rm in}(t),\omega_{\rm out}(t))$. \\

LANs\_complete &
Instantaneous longitudes of ascending node of the inner and outer orbits,
$(\Omega_{\rm in}(t),\Omega_{\rm out}(t))$. \\



tAnos\_complete &
Instantaneous true anomalies of the inner and outer orbits,
$(\nu_{\rm in}(t),\nu_{\rm out}(t))$, describing the orbital phases.
These quantities may become undefined when the orbital elements become degenerate
or the system becomes unstable. \\






\bottomrule
\end{tabularx}
\end{table*}

\paragraph{Dynamic summary of the prefix} 
\label{app:dynamical_summary}
Dynamical summary statistics extracted from each trajectory channel over the observed prefix (Table ~\ref{tab:dynamical_summary}). For a channel $x(t)$ with prefix length $T$, the same set of statistics is computed independently for all 15 orbital channels.
Dynamic summary statistics which are used as features for some tasks.

\begin{table}[!htbp]
\centering
\caption{
Dynamic summary statistics which are used as features for some tasks.
}
\label{tab:dynamical_summary}
\small
\begin{tabular}{l l}
\toprule
Statistic & Definition \\
\midrule
Mean 
& $\displaystyle \frac{1}{T}\sum_{t=1}^{T} x(t)$ \\[4pt]

Standard deviation 
& $\displaystyle \sqrt{\frac{1}{T}\sum_{t=1}^{T}\left(x(t)-\bar{x}\right)^2}$ \\[6pt]

Minimum 
& $\displaystyle \min_{1\leq t\leq T} x(t)$ \\[4pt]

Maximum 
& $\displaystyle \max_{1\leq t\leq T} x(t)$ \\[4pt]

Range 
& $\displaystyle \max_t x(t)-\min_t x(t)$ \\[4pt]

First value 
& $x(1)$ \\

Last value 
& $x(T)$ \\

Net change ($\Delta$) 
& $x(T)-x(1)$ \\

Linear slope 
& Slope coefficient obtained by least-squares linear regression of $x(t)$ on $t$ \\
\bottomrule
\end{tabular}
\end{table}

\subsection{Data Splits}
\label{app:data_splits}

To prevent information leakage between closely related initial-condition
variations, all train--validation--test partitions are constructed at the
underlying-ensemble level. In particular, individual systems generated from the same
ensemble of systems are never assigned to different splits.

\paragraph{Stability prediction splits.}
The initial-condition-based prediction tasks and the short-term-prefix-based
stability prediction tasks share the same ensemble-level split. After removing
incomplete systems, the dataset contains 4,998 ensembles of systems and 159,936
individual systems, with 32 initial-condition variations per ensemble of systems. We use a
$70\%/15\%/15\%$ train--validation--test split with random seed 42, resulting
in 3,498 training ensembles of systems, 750 validation ensembles of systems, and 750 test ensembles of systems.

For ensemble-level prediction, each ensemble of systems constitutes one sample. For
individual-level stability classification, the individual trajectories are
treated as separate samples only after the ensemble-level split has been
determined. The resulting individual system counts are therefore 111,936,
24,000, and 24,000 for the training, validation, and test sets,
respectively. Varying the observed prefix length or the number of
initial-condition variations used by a model does not alter the underlying system assignment.

\begin{table}[htbp]
\centering
\small
\caption{
Ensemble-level split used for the initial-condition-based and
short-term-prefix-based stability prediction experiments. All 32
initial-condition variations associated with an ensemble of systems are kept in the same
partition.
}
\label{tab:stability_split}
\begin{tabular}{lrr}
\toprule
Split & Ensembles & Individuals \\
\midrule
Train      & 3,498 & 111,936 \\
Validation &   750 &  24,000 \\
Test       &   750 &  24,000 \\
\bottomrule
\end{tabular}
\end{table}

\paragraph{Trajectory forecasting split.}
Trajectory forecasting requires a separate split because a trajectory must
contain a sufficiently long valid segment to construct a complete
context--prediction window. Under our default forecasting protocol, each
example contains a 512-step context followed by a 128-step prediction
horizon. We therefore first retain only trajectories for which at least one
complete 640-step forecasting window can be constructed.

After this completion-aware filtering, the forecasting subset contains
64,597 individual systems from 3,847 ensembles of systems. These eligible systems are then
independently divided using the same $70\%/15\%/15\%$ proportions and
random seed 42, yielding 2,692 training ensembles of systems, 577 validation ensembles of systems,
and 578 test ensembles of systems. The split remains strictly ensemble-disjoint, although
not every eligible ensemble of system necessarily retains all 32 individual system after
forecasting-window filtering.

\begin{table}[htbp]
\centering
\small
\caption{
Ensemble-level split used for trajectory forecasting after completion-aware
filtering under the $512\!\rightarrow\!128$ forecasting protocol.
}
\label{tab:forecasting_split}
\begin{tabular}{lrr}
\toprule
Split & Ensembles & Individuals \\
\midrule
Train      & 2,692 & 44,988 \\
Validation &   577 &  9,961 \\
Test       &   578 &  9,648 \\
\midrule
Total      & 3,847 & 64,597 \\
\bottomrule
\end{tabular}
\end{table}

Because forecasting windows extracted from the same trajectory can strongly
overlap, we subsample candidate windows during training and evaluation.
During training, at most eight windows are randomly sampled from each
trajectory per epoch. Validation and one-shot forecasting evaluation use up
to four fixed windows per trajectory, while long-horizon rolling evaluation
uses up to two starting points per trajectory. Feature normalization
statistics are estimated exclusively from finite, pre-instability
observations in the training split.

\subsection{Evaluation Metric Definitions}
\label{app:evaluation_metrics}

We evaluate all models on held-out systems. Unless otherwise
specified, reported metrics are computed on the test split after model and
hyperparameter selection using only the training and validation splits.

\paragraph{Binary classification.}
For a binary target $y_i\in\{0,1\}$, let $\hat{p}_i$ denote the predicted
probability of the positive class and
$\hat{y}_i=\mathbb{I}[\hat{p}_i\geq\tau]$ the corresponding binary prediction.
Accuracy and F1 are defined as
\begin{equation}
    \mathrm{Accuracy}
    =
    \frac{\mathrm{TP}+\mathrm{TN}}
         {\mathrm{TP}+\mathrm{TN}+\mathrm{FP}+\mathrm{FN}},
\end{equation}
and
\begin{equation}
    \mathrm{F1}
    =
    \frac{2\,\mathrm{Precision}\,\mathrm{Recall}}
         {\mathrm{Precision}+\mathrm{Recall}},
    \qquad
    \mathrm{Precision}=\frac{\mathrm{TP}}{\mathrm{TP}+\mathrm{FP}},
    \quad
    \mathrm{Recall}=\frac{\mathrm{TP}}{\mathrm{TP}+\mathrm{FN}}.
\end{equation}
When a threshold $\tau$ is optimized, it
is selected exclusively on the validation split by maximizing F1 and is then
kept fixed for test evaluation. We additionally report AUROC and AUPRC, which
are threshold-independent and are computed from $\hat{p}_i$. The positive
class is explicitly indicated in each result table; stability is treated as
the positive class in tables reporting ``stable F1.'' Because F1 and AUPRC
depend on the positive-class convention, stable-positive and
unstable-positive values are not directly interchangeable.

For individual-level classification, each individual perturbation is treated
as one example. For ensemble-level evaluation, probabilities from the $V_i$
perturbations of ensemble of systems $i$ are averaged:
\begin{equation}
    \hat{p}^{\mathrm{ensemble}}_i
    =
    \frac{1}{V_i}\sum_{j=1}^{V_i}\hat{p}_{ij}.
\end{equation}
The corresponding ensemble target is the observed fraction of stable
perturbations. In the complete ensemble experiments, $V_i=32$.

\paragraph{Regression.}
For targets $y_i$ and predictions $\hat{y}_i$, we report
\begin{align}
    \mathrm{MAE}
    &= \frac{1}{N}\sum_{i=1}^{N}
       \left|\hat{y}_i-y_i\right|,\\
    \mathrm{MSE}
    &= \frac{1}{N}\sum_{i=1}^{N}
       \left(\hat{y}_i-y_i\right)^2,\\
    \mathrm{RMSE}
    &= \sqrt{\frac{1}{N}\sum_{i=1}^{N}
       \left(\hat{y}_i-y_i\right)^2},\\
    R^2
    &=1-
      \frac{\sum_i(\hat{y}_i-y_i)^2}
           {\sum_i(y_i-\bar{y})^2}.
\end{align}
Pearson correlation measures linear agreement between predictions and
targets, while Spearman correlation, where reported, measures agreement
between their ranks. Predictions of bounded quantities such as stable
fractions are clipped to their valid range $[0,1]$ before evaluation.

\paragraph{Survival prediction.}
Let $T_{ij}$ be the event time of perturbation $j$ in ensemble of system $i$, and let
\begin{equation}
    Y_{ijk}=\mathbb{I}[T_{ij} \geq t_k]
\end{equation}
be its empirical survival indicator at evaluation horizon $t_k$. Given a
predicted survival probability $\hat{S}_{ij}(t_k)$, the predicted
ensemble-level survival fraction is
\begin{equation}
    \hat{S}^{\mathrm{ensemble}}_i(t_k)
    =
    \frac{1}{V_i}\sum_{j=1}^{V_i}\hat{S}_{ij}(t_k).
\end{equation}
The horizon-wise ensemble MAE is
\begin{equation}
    \mathrm{MAE}(t_k)
    =
    \frac{1}{N}\sum_{i=1}^{N}
    \left|
      \hat{S}^{\mathrm{ensemble}}_i(t_k)-S^{\mathrm{ensemble}}_i(t_k)
    \right|,
\end{equation}
and the overall survival MAE is its mean across evaluation horizons.

The Brier score at horizon $t_k$ is evaluated at the perturbation level:
\begin{equation}
    \mathrm{BS}(t_k)
    =
    \frac{1}{\sum_i V_i}
    \sum_{i=1}^{N}\sum_{j=1}^{V_i}
    \left(
      \hat{S}_{ij}(t_k)-Y_{ijk}
    \right)^2.
\end{equation}
Our discrete integrated Brier score is
\begin{equation}
    \mathrm{IBS}
    =
    \frac{1}{K}\sum_{k=1}^{K}\mathrm{BS}(t_k).
\end{equation}
The reported experiments use $K=20$ logarithmically spaced time bins between
$t=5$ and $t=10{,}000$. This is an equally weighted discrete approximation:
each ensemble of systems,  each perturbations, and each time bins receive equal weight. Event indicators
are observed over the complete simulation interval; individual systems that remain
bound at the terminal horizon are treated as surviving through the evaluation
window. Consequently, the reported IBS does not apply inverse-probability-of-censoring weighting (IPCW).

\paragraph{Trajectory forecasting.}
Continuous trajectory channels are standardized using statistics estimated
from the training split only:
\begin{equation}
    z_{itc}
    =
    \frac{x_{itc}-\mu^{\mathrm{train}}_c}
         {\sigma^{\mathrm{train}}_c}.
\end{equation}
We compute MSE, MAE, and RMSE in this standardized feature space. Scalar forecasting metrics are ensemble-balanced: errors are first averaged over forecast horizons and channels within each prediction window, then over windows from the same individual system, and finally over ensemble of systems. This prevents systems contributing different numbers of windows from receiving different weights.

Pearson correlation is computed within prediction windows and aggregated using the same ensemble-balanced hierarchy. Horizon-wise errors are obtained by
holding the forecast step fixed and averaging over channels, individual systems, and
ensembles of systems. For autoregressive rollouts, these metrics are reported separately
for successive forecast blocks to characterize accumulated error.

Angular variables are represented by sine--cosine pairs. In addition to their
standardized forecasting errors, we measure violation of the unit-circle
constraint using
\begin{equation}
    E_{\mathrm{circle}}
    =
    \frac{1}{|\mathcal{A}|}
    \sum_{a\in\mathcal{A}}
    \left|
      \hat{s}_a^2+\hat{c}_a^2-1
    \right|,
\end{equation}
where $(\hat{s}_a,\hat{c}_a)$ is a predicted sine--cosine pair and
$\mathcal{A}$ contains all predicted angular pairs, horizons, and evaluated
examples. Lower values indicate that the predicted representation remains
closer to the unit circle. This metric measures geometric consistency of the
angular representation rather than angular displacement from the target.

\section{Full Results}
\label{app:result}


\subsection{Task A: Stability Prediction from Initial Conditions}
\begin{table}[!htbp]
\centering
\caption{Binary stability classification from initial physical conditions. Stability is treated as the positive class.
}
\label{tab:task_a_classification_result}
\small
\begin{tabular}{lcccccc}
\toprule
Method & Accuracy & Precision & Recall & Stable F1 & AUROC & AUPRC \\
\midrule
Mardling--Aarseth criterion~\cite{mardling2001tidal} & 0.6187 & 0.2875 & \textbf{0.9496} & 0.4414 & 0.7529 & 0.2810 \\
Vynatheya et al. criterion~\cite{vynatheya2022algebraic} & 0.6600 & 0.3046 & 0.8908 & 0.4540 & 0.7536 & 0.2887 \\
Vynatheya et al. MLP classifier ~\cite{vynatheya2022algebraic,vynatheya2023quadruple} & 0.7267 & 0.3547 & 0.8824	& 0.5060 & \textbf{0.9086} & \textbf{0.7184} \\
MLP          & 0.7507 & 0.3629 & 0.7563 & 0.4905 & 0.8416 & 0.5139 \\
XGBoost      & \textbf{0.8053} & \textbf{0.4308} & 0.7059 & \textbf{0.5350} & 0.8525 & 0.5089 \\
\bottomrule
\end{tabular}
\end{table}

\begin{table}[!htbp]
\centering
\caption{Stable-fraction regression from initial physical conditions.
}
\label{tab:task_a_regression_result}
\small
\begin{tabular}{lcccc}
\toprule
Method & MSE & RMSE & MAE & $R^2$ \\
\midrule
Mardling-Aarseth criterion ~\cite{mardling2001tidal} & 0.2488	& 0.4988 & 0.3073	& -0.5653 \\
Vynatheya et al. criterion~\cite{vynatheya2022algebraic} & 0.2268 & 0.4763 & 0.2853 & 	-0.4269 \\
Vynatheya et al. MLP classifier ~\cite{vynatheya2022algebraic,vynatheya2023quadruple} & 0.1056 & 0.3249 & \textbf{0.2176} & 0.3359 \\
MLP     & 0.0901 & 0.3001 & 0.2247 & 0.4335 \\
XGBoost & \textbf{0.0884} & \textbf{0.2972} & 0.2330 & \textbf{0.4442} \\
\bottomrule
\end{tabular}
\end{table}

\subsection{Task B: Stability Prediction from Early Dynamics}
Results are shown in Table ~\ref{tab:task_b_result}.
\begin{table}[!htbp]
\centering
\caption{
Performance of different trajectory-prefix representations.
Individual-level predictions are additionally aggregated over the 32 variations
of each ensemble of system to evaluate ensemble-level stable-fraction prediction.
Stability is treated as the positive class.}
\label{tab:task_b_result}
\small
\setlength{\tabcolsep}{3.5pt}
\resizebox{\linewidth}{!}{
\begin{tabular}{c l l cccc ccc}
\toprule
& & &
\multicolumn{4}{c}{Individual-level classification} &
\multicolumn{3}{c}{Ensemble-level prediction} \\
\cmidrule(lr){4-7}
\cmidrule(lr){8-10}
Prefix length&
Representation &
Model &
Acc. &
Stable F1 &
AUROC &
AUPRC &
MAE &
MSE &
$R^2$ \\
\midrule

\multirow{6}{*}{96}
& PANDA encoder        & Linear  & 0.8640 & 0.8070 & 0.9467 & 0.8911 & 0.1031 & 0.0256 & 0.8390 \\
& Chronos-2 encoder    & Linear  & 0.8623 & 0.8067 & 0.9531 & 0.9083 & 0.1112 & 0.0281 & 0.8230 \\
& Dynamical summaries  & XGBoost &
\textbf{0.9000} &
\textbf{0.8404} &
\textbf{0.9649} &
\textbf{0.9319} &
\textbf{0.0703} &
\textbf{0.0153} &
\textbf{0.9040} \\
& \texttt{orbital15}         & SVM     & 0.7493 & 0.5800 & 0.8135 & 0.6166 & 0.2373 & 0.0927 & 0.4165 \\
& \texttt{orbital15}         & CNN     & 0.7937 & 0.6092 & 0.8888 & 0.7526 & 0.1759 & 0.0786 & 0.5056 \\
& \texttt{orbital15}         & TCN     & 0.8167 & 0.7402 & 0.8971 & 0.7523 & 0.1570 & 0.0581 & 0.6347 \\

\midrule

\multirow{6}{*}{512}
& PANDA encoder        & Linear  & 0.9217 & 0.8845 & 0.9786 & 0.9550 & 0.0534 & 0.0071 & 0.9554 \\
& Chronos-2 encoder    & Linear  & 0.9209 & 0.8848 & 0.9791 & 0.9552 & 0.0614 & 0.0087 & 0.9452 \\
& Dynamical summaries  & XGBoost &
\textbf{0.9329} &
\textbf{0.8960} &
\textbf{0.9851} &
\textbf{0.9673} &
\textbf{0.0398} &
\textbf{0.0051} &
\textbf{0.9676} \\
& Raw \texttt{orbital15}         & SVM     & 0.7590 & 0.5918 & 0.8282 & 0.6232 & 0.2184 & 0.0818 & 0.4855 \\
& Raw \texttt{orbital15}        & CNN     & 0.8839 & 0.8205 & 0.9470 & 0.8543 & 0.0985 & 0.0278 & 0.8249 \\
& Raw \texttt{orbital15}         & TCN     & 0.8973 & 0.8411 & 0.9593 & 0.8900 & 0.0765 & 0.0177 & 0.8886 \\

\bottomrule
\end{tabular}
}
\end{table}

\paragraph{Task B ablation: At-risk evaluation.}
We mainly carry out two ablations at both individual and ensemble level to remove the systems that have already got unstable within the prefix. Results are shown in Table ~\ref{tab: system-level at-risk} and Table ~\ref{tab:trajectory_at_risk}.
\begin{table}[!htbp]
\centering
\caption{
At-risk ablation for ensemble-level stability prediction.
Evaluation is restricted to systems that remain bound through the specified
time $t_{\mathrm{risk}}$. All models use XGBoost.
Lower is better for both metrics.
}
\label{tab: system-level at-risk}
\small
\begin{tabular}{llccc}
\toprule
At-risk cohort
& Representation
& Ensembles
& MAE 
& MSE  \\
\midrule

\multirow{2}{*}{$Prefix>96$}
& Initial conditions
& 257
& 0.3432
& 0.1551 \\
& Dynamical summary, prefix 96
& 257
& \textbf{0.1103}
& \textbf{0.0269} \\

\midrule

\multirow{3}{*}{$Prefix>512$}
& Initial conditions
& 174
& 0.3985
& 0.1940 \\
& Dynamical summaries, prefix 96
& 174
& 0.0897
& 0.0187 \\
& Dynamical summaries, prefix 512
& 174
& \textbf{0.0520}
& \textbf{0.0079} \\

\bottomrule
\end{tabular}
\vspace{2pt}
\begin{minipage}{0.98\linewidth}
\footnotesize
At-risk cohorts include only individual systems that remain bound up to the specified time.
\end{minipage}
\end{table}

\begin{table}[!htbp]
\centering
\caption{
At-risk individual-level stability classification.
Only individual systems that remain bound through the specified at-risk time $t_{\mathrm{risk}}$ are retained. Stability is treated as the positive class.
}
\label{tab:trajectory_at_risk}
\small
\setlength{\tabcolsep}{2.5pt}
\begin{tabular}{c c l l r r c c c c}
\toprule
At-risk cohort
& Prefix
& Representation
& Predictor
& Individuals
& Stable \%
& Acc.
& Stable F1
& AUROC
& AUPRC \\
\midrule

\multirow{3}{*}{$Prefix > 96$}
& \multirow{3}{*}{96}
& PANDA 
& Linear probe
& \multirow{3}{*}{14,192}
& \multirow{3}{*}{54.04}
& 0.7709
& 0.8075
& 0.8716
& 0.8923 \\

&
& Chronos-2
& Linear probe
&
&
& 0.7672
& 0.8067
& 0.8848
& 0.9088 \\

&
& Dynamical summaries
& XGBoost
&
&
& \textbf{0.8309}
& \textbf{0.8404}
& \textbf{0.9123}
& \textbf{0.9320} \\

\midrule

\multirow{6}{*}{$Prefix > 512$}
& \multirow{3}{*}{96}
& PANDA 
& Linear probe
& \multirow{6}{*}{10,189}
& \multirow{6}{*}{75.27}
& 0.7877
& 0.8631
& 0.8172
& 0.9301 \\

&
& Chronos-2 
& Linear probe
&
&
& 0.7885
& \textbf{0.8649}
& 0.8341
& 0.9398 \\

&
& Dynamical summaries
& XGBoost
&
&
& \textbf{0.7999}
& 0.8611
& \textbf{0.8655}
& \textbf{0.9537} \\

\cmidrule(lr){2-4}
\cmidrule(lr){7-10}

&
\multirow{3}{*}{512}
& PANDA 
& Linear probe
&
&
& 0.8180
& 0.8859
& 0.8787
& 0.9578 \\

&
& Chronos-2
& Linear probe
&
&
& 0.8147
& 0.8854
& 0.8781
& 0.9573 \\

&
& Dynamical summaries
& XGBoost
&
&
& \textbf{0.8421}
& \textbf{0.8961}
& \textbf{0.9040}
& \textbf{0.9674} \\

\bottomrule
\end{tabular}

\vspace{2pt}
\begin{minipage}{0.98\linewidth}
\footnotesize
For an at-risk time $t_P$, evaluation retains only individual systems satisfying $T>t_P$ and predicts whether they remain stable through $H=10{,}000$.
The strict inequality excludes individual systems that become unbound exactly at the observation endpoint.
All classification metrics use a fixed probability threshold of $0.5$,
with stability treated as the positive class.
Stable \% denotes the positive-class prevalence.
\end{minipage}
\end{table}

\subsection{Task C: Multi-horizon Survival Prediction}
Results are shown in Table ~\ref{tab:task_c_survival}.

\begin{table}[!htbp]
\centering
\caption{
Multi-horizon survival prediction performance.
All learned representations use the same MLP hazard head.
}
\label{tab:task_c_survival}
\small
\setlength{\tabcolsep}{6pt}
\renewcommand{\arraystretch}{1.08}

\begin{tabularx}{\linewidth}{l X ccc}
\toprule
Method
& Aggregation
& Overall MAE 
& Final MAE 
& IBS  \\
\midrule

\multicolumn{5}{l}{\textit{Baselines}} \\
Train-mean
& --
& 0.3099
& 0.3510
& 0.1315 \\

Raw \texttt{orbital15} CNN
& --
& 0.0910
& 0.1728
& 0.0752 \\

Raw \texttt{orbital15} Dynamical summaries
& --
& 0.0670
& 0.1475
& 0.0632 \\

Raw \texttt{orbital15} TCN
& --
& 0.0662
& 0.1564
& 0.0606 \\

\midrule
\multicolumn{5}{l}{\textit{Pretrained time-series representations}} \\

\multirow{3}{*}{PANDA}
& Mean/std pooling
& 0.0602
& 0.1035
& 0.0550 \\

& Channel-attention pooling
& 0.0459
& 0.0892
& 0.0486 \\

& Direct
& 0.0427
& \textbf{0.0819}
& 0.0459 \\

\addlinespace[2pt]

\multirow{3}{*}{Chronos-2}
& Mean/std pooling
& 0.0476
& 0.0939
& 0.0480 \\

& Channel-attention pooling
& 0.0414
& 0.0887
& 0.0426 \\

& Direct
& \textbf{0.0376}
& 0.0826
& \textbf{0.0416} \\

\bottomrule
\end{tabularx}

\vspace{2pt}
\begin{minipage}{0.98\linewidth}
\footnotesize
Overall MAE is averaged over all 20 prediction horizons,
whereas Final MAE evaluates only the final horizon.
IBS denotes the integrated Brier score.
\end{minipage}

\end{table}

\subsection{Long-horizon Trajectory Forecasting}
\begin{table}[!htbp]
\centering
\caption{Trajectory forecasting performance. }
\label{tab:trajectory_forecasting}
\small
\begin{tabular}{llcccc}
\toprule
Model & Setting & MSE (z) & MAE (z) & Pearson & Unit-circle Error \\
\midrule
Chronos-2 ~\cite{ansari2025chronos2}   & Zero-shot  & 0.2331 & \textbf{0.2282} & 0.8580 & \textbf{0.2265} \\
PANDA ~\cite{lai2026panda}       & Zero-shot  & 0.2217 & 0.2371 & 0.8625 & 0.2862 \\
iTransformer ~\cite{liu2024itransformer} & Supervised & \textbf{0.2123} & 0.2429 & \textbf{0.8673} & 0.3152 \\
TOTO2 ~\cite{khwaja2026toto}       & Zero-shot  & 0.2262 & 0.2303 & 0.8601 & 0.2310 \\
TiRex2 ~\cite{podest2026tirex}      & Zero-shot  & 0.2422 & 0.2408 & 0.8508 & 0.2277 \\
\bottomrule
\end{tabular}
\end{table}

\begin{figure}[!htbp]
    \centering
    \includegraphics[width=\linewidth]{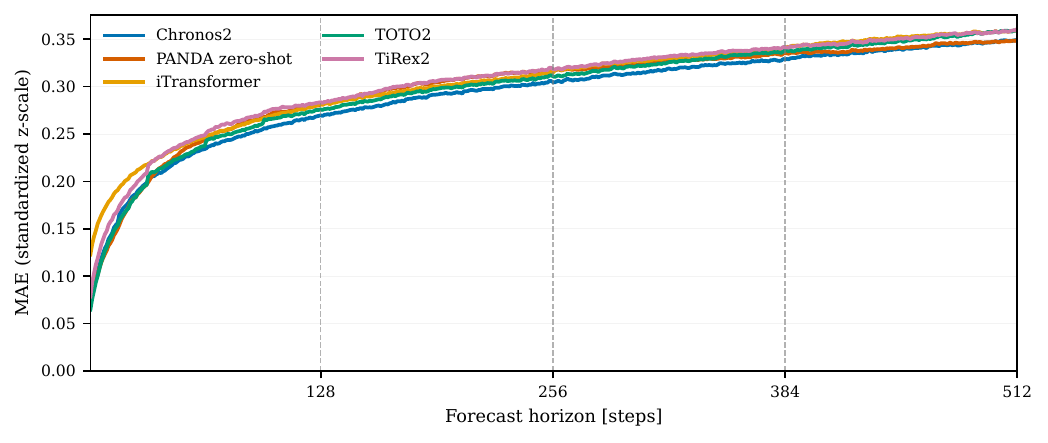}
    \caption{Error Accumulation during Autoregressive Long-Horizon Rollouts.}
    \label{fig:rolling_horizon_mae}
\end{figure}

\end{document}

%% file: checklist.tex
\section*{NeurIPS Paper Checklist}

\begin{enumerate}

\item {\bf Claims}
    \item[] Question: Do the main claims made in the abstract and introduction accurately reflect the paper's contributions and scope?
    \item[] Answer: \answerYes{} 
    \item[] Justification: Our claims in the abstract and introduction are accurate reflections of the paper's contributions and scope.

\item {\bf Limitations}
    \item[] Question: Does the paper discuss the limitations of the work performed by the authors?
    \item[] Answer: \answerYes{} 
    \item[] Justification: We discuss the limitations of current ML approaches in predicting longterm dynamics in the final section.

\item {\bf Theory assumptions and proofs}
    \item[] Question: For each theoretical result, does the paper provide the full set of assumptions and a complete (and correct) proof?
    \item[] Answer: \answerNA{} 
    \item[] Justification: We do not present any new theoretical contributions to machine learning in this work. However, the adopted simulation codes are based on well-established astrophysical theories and assumptions.

    \item {\bf Experimental result reproducibility}
    \item[] Question: Does the paper fully disclose all the information needed to reproduce the main experimental results of the paper to the extent that it affects the main claims and/or conclusions of the paper (regardless of whether the code and data are provided or not)?
    \item[] Answer: \answerYes{} 
    \item[] Justification: We fully disclose the implementation details of our key modeling components, training, and inference procedures in the paper and appendices. 

\item {\bf Open access to data and code}
    \item[] Question: Does the paper provide open access to the data and code, with sufficient instructions to faithfully reproduce the main experimental results, as described in supplemental material?
    \item[] Answer: \answerYes{} 
    \item[] Justification: The code and dataset are publicly available at the links provided in the paper and repository.

\item {\bf Experimental setting/details}
    \item[] Question: Does the paper specify all the training and test details (e.g., data splits, hyperparameters, how they were chosen, type of optimizer) necessary to understand the results?
    \item[] Answer: \answerYes{} 
    \item[] Justification: We have provided implementation details covering the necessary training and test details to understand the results.

\item {\bf Experiment statistical significance}
    \item[] Question: Does the paper report error bars suitably and correctly defined or other appropriate information about the statistical significance of the experiments?
    \item[] Answer: \answerNo{} 
    \item[] Justification: We report point estimates on a fixed train–validation–test split, without error bars, confidence intervals, or statistical significance tests. As this is a workshop paper presenting our ongoing work, we plan to quantify variability across repeated runs in an extended version of this work.

\item {\bf Experiments compute resources}
    \item[] Question: For each experiment, does the paper provide sufficient information on the computer resources (type of compute workers, memory, time of execution) needed to reproduce the experiments?
    \item[] Answer: \answerYes{} 
    \item[] Justification: Detailed information is provided in the public GitHub repository linked in the paper.
    
\item {\bf Code of ethics}
    \item[] Question: Does the research conducted in the paper conform, in every respect, with the NeurIPS Code of Ethics \url{https://neurips.cc/public/EthicsGuidelines}?
    \item[] Answer: \answerYes{} 
    \item[] Justification: This research conforms with the NeurIPS Code of Ethics.

\item {\bf Broader impacts}
    \item[] Question: Does the paper discuss both potential positive societal impacts and negative societal impacts of the work performed?
    \item[] Answer: \answerNA{} 
    \item[] Justification: We do not foresee any significant societal impact of our work as the main application is within astronomical data. 
    
\item {\bf Safeguards}
    \item[] Question: Does the paper describe safeguards that have been put in place for responsible release of data or models that have a high risk for misuse (e.g., pre-trained language models, image generators, or scraped datasets)?
    \item[] Answer: \answerNA{} 
    \item[] Justification: This paper does not pose the described risks. 

\item {\bf Licenses for existing assets}
    \item[] Question: Are the creators or original owners of assets (e.g., code, data, models), used in the paper, properly credited and are the license and terms of use explicitly mentioned and properly respected?
    \item[] Answer: \answerYes{} 
    \item[] Justification: The code, data, and models used in this paper has been properly credited with corresponding license and terms of use.

\item {\bf New assets}
    \item[] Question: Are new assets introduced in the paper well documented and is the documentation provided alongside the assets?
    \item[] Answer: \answerYes{} 
    \item[] Justification: Dataset details are included in the main paper and appendix.

\item {\bf Crowdsourcing and research with human subjects}
    \item[] Question: For crowdsourcing experiments and research with human subjects, does the paper include the full text of instructions given to participants and screenshots, if applicable, as well as details about compensation (if any)? 
    \item[] Answer: \answerNA{} 
    \item[] Justification: This research does not involve crowdsourcing nor research with human subjects.

\item {\bf Institutional review board (IRB) approvals or equivalent for research with human subjects}
    \item[] Question: Does the paper describe potential risks incurred by study participants, whether such risks were disclosed to the subjects, and whether Institutional Review Board (IRB) approvals (or an equivalent approval/review based on the requirements of your country or institution) were obtained?
    \item[] Answer: \answerNA{} 
    \item[] Justification: This research does not involve crowdsourcing nor research with human subjects.

\item {\bf Declaration of LLM usage}
    \item[] Question: Does the paper describe the usage of LLMs if it is an important, original, or non-standard component of the core methods in this research? Note that if the LLM is used only for writing, editing, or formatting purposes and does \emph{not} impact the core methodology, scientific rigor, or originality of the research, declaration is not required.
    \item[] Answer: \answerNA{} 
    \item[] Justification: We do not use LLMs for any core design of the proposed method, but only for minor writing editing.

\end{enumerate}